\documentclass[conference,a4paper]{IEEEtran}
\IEEEoverridecommandlockouts
\usepackage{cite}
\usepackage{amsmath,amssymb,amsfonts}
\usepackage{algorithmic}
\usepackage{graphicx}
\usepackage{textcomp}
\usepackage{xcolor}

\usepackage{breqn}
\usepackage{siunitx}
\usepackage{bm}
\usepackage{hyperref}
\usepackage{booktabs}
\usepackage{multirow}
\usepackage{threeparttable}
\usepackage{mathtools}
\usepackage{doi}

\def\BibTeX{{\rm B\kern-.05em{\sc i\kern-.025em b}\kern-.08em
    T\kern-.1667em\lower.7ex\hbox{E}\kern-.125emX}}
\begin{document}

\sisetup{mode=match}

\title{Event-Based Optical Flow Constraint\\ via Time-Sliced Plane Fitting
\thanks{This work was supported by JSPS KAKENHI Grant Number JP24K14996.}%
\thanks{\copyright~2026 IEEE. Personal use of this material is permitted. Permission from IEEE must be obtained for all other uses, in any current or future media, including reprinting/republishing this material for advertising or promotional purposes, creating new collective works, for resale or redistribution to servers or lists, or reuse of any copyrighted component of this work in other works. Accepted for publication at IEEE VCIP 2026.}
}
\author{
\IEEEauthorblockN{
Masaki Koike\IEEEauthorrefmark{2}, 
Yusuke Kameda\IEEEauthorrefmark{3},
Shunichi Sato\IEEEauthorrefmark{2}, and 
Takayuki Hamamoto\IEEEauthorrefmark{2}
}
\IEEEauthorblockA{\IEEEauthorrefmark{2}\textit{Graduate School of Engineering, Tokyo University of Science,} Tokyo, Japan}
\IEEEauthorblockA{\IEEEauthorrefmark{3}\textit{Faculty of Science and Technology, Sophia University,} Tokyo, Japan}
\IEEEauthorblockA{Email: 4326527@ed.tus.ac.jp, kameda@sophia.ac.jp, \{sato.shunichi, hamamoto\}@rs.tus.ac.jp}
\IEEEauthorblockA{ORCID: \href{https://orcid.org/0000-0001-8503-4098}{0000-0001-8503-4098} (Y. Kameda), \href{https://orcid.org/0000-0001-8246-8325}{0000-0001-8246-8325} (T. Hamamoto)}
}

\maketitle

\begin{abstract}
Event-based optical flow estimation using spatiotemporal planar approximation inherently assumes constant velocity and tends to average out temporal velocity variations. In this paper, we propose a flow estimation method that extends this planar approximation along the temporal axis. The resulting event-based optical flow constraint allows existing variational optical flow formulations to model temporal motion variation. Specifically, we partition the space-time volume into sub-intervals and apply a two-step planar approximation to track object edges consistently. Moreover, a spatial compensation mechanism and a temporal regularization term are incorporated to compensate for edge displacements and temporally stabilize the flow estimation, respectively. The experimental results show that modeling temporal motion variation improves flow estimation accuracy over conventional methods.
\end{abstract}


\section{Introduction}
An event camera is a bio-inspired vision sensor that asynchronously detects per-pixel brightness changes, outputting events $\bm{e}$ with timestamp $t$, pixel coordinates $\bm{x} = (x, y)^\top$, and the polarity of the brightness change $p\in \{+,-\}$. 
Due to their high temporal resolution on the order of \si{\us}, high dynamic range (HDR) and low power consumption~\cite{9138762}, they are increasingly utilized in applications such as object detection~\cite{11097744}, HDR video reconstruction~\cite{11444930}, and edge systems~\cite{8836544,10.1145/3643832.3661415}.

Optical flow estimation determines apparent displacements between time points, and is applied in fields such as video compression~\cite{9165938} and object detection~\cite{7917999}. As event cameras fundamentally differ from conventional frames, existing frame-based approaches are not directly applicable.
To estimate flow without frame conversion, methods such as contrast maximization~\cite{10517639} and spatiotemporal planar approximation of the event point cloud~\cite{10.1117/1.JEI.34.4.043024} have been proposed.

A previous study~\cite{10.1117/1.JEI.34.4.043024} estimates optical flow via planar approximation of the event point cloud, but this formulation inherently enforces a velocity constancy assumption within the temporal estimation window, yielding only an average velocity field.
This limitation may be particularly problematic in practical scenarios. Even when objects move at constant velocity in 3D space, their projected motion on the image plane can exhibit apparent acceleration---for instance, during camera translation, points near the Focus of Expansion (FOE) move slowly while peripheral points move faster.
Such spatially varying apparent acceleration cannot be captured under a constant-velocity assumption, motivating the need for an acceleration-aware estimation method.

In this paper, we partition the space-time volume in the temporal direction and estimate flow for each sub-interval, enabling the model to account for apparent acceleration during flow estimation arising from both object motion and perspective projection.

This paper is an extended version of a non-peer-reviewed technical report~\cite{pcsj2025}.

\section{Related Work}
Event-based optical flow estimation methods can be broadly categorized into learning-based and model-based approaches~\cite{9138762}. 
Learning-based methods (e.g., EV-FlowNet~\cite{zhu2018ev}, E-RAFT~\cite{gehrig2021raft}) achieve high accuracy but require extensive training and lack geometric interpretability.
Among model-based methods, the contrast maximization framework~\cite{10517639} is widely used, which estimates motion parameters by maximizing the sharpness of motion-compensated event images.

Alternatively, another class of model-based approaches~\cite{10.1117/1.JEI.34.4.043024} explicitly leverages the local geometric structure of events in space-time. By extracting spatiotemporal gradients directly from event surfaces, this approach uniquely allows the integration of classical, mathematically well-founded optical flow frameworks (such as the Horn--Schunck method~\cite{horn1981determining}) without requiring pseudo-frame generation.
\begin{figure}[t]
    \centering
    \includegraphics[width=0.9\linewidth]{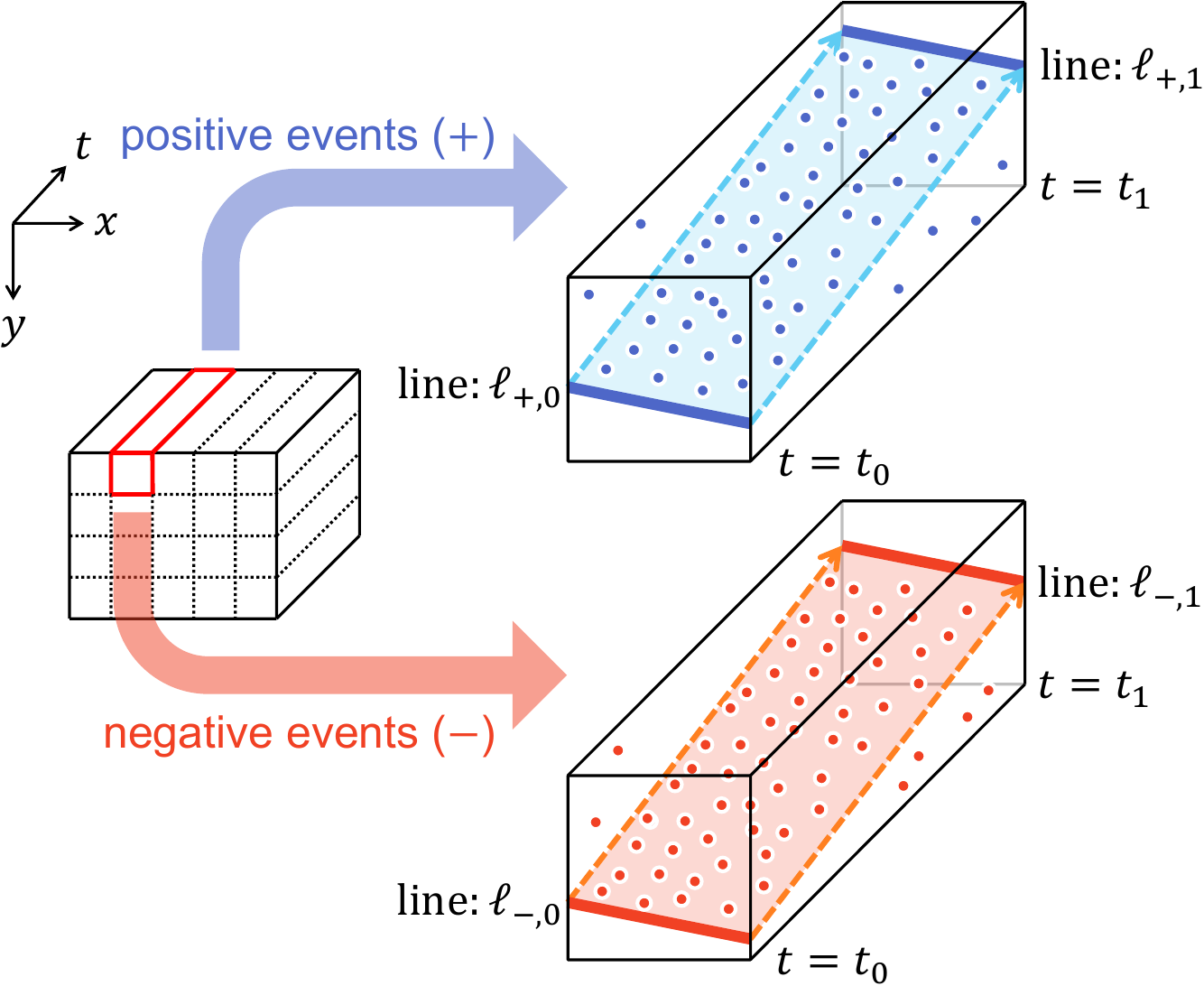}
    \caption{Concept of optical flow estimation based on spatiotemporal planar approximation. In subsequent figures, the distinction between positive and negative polarities is omitted, and only one polarity is illustrated.}
    \label{fig:conventional_method}
\end{figure}
Specifically, a previous study~\cite{10.1117/1.JEI.34.4.043024} assumes that when an object edge moves, the resulting event point cloud forms a surface in space-time along its trajectory. Optical flow is then estimated by approximating this surface as a single plane via RANSAC (Fig.~\ref{fig:conventional_method}), applied to each spatially partitioned sub-region. However, this formulation shares the constant-velocity assumption within the temporal estimation window. The proposed method addresses this limitation by extending the planar approximation along the temporal axis.

\section{Preliminaries} \label{sec:preliminaries}

\begin{figure}
    \centering
    \includegraphics[width=0.9\linewidth]{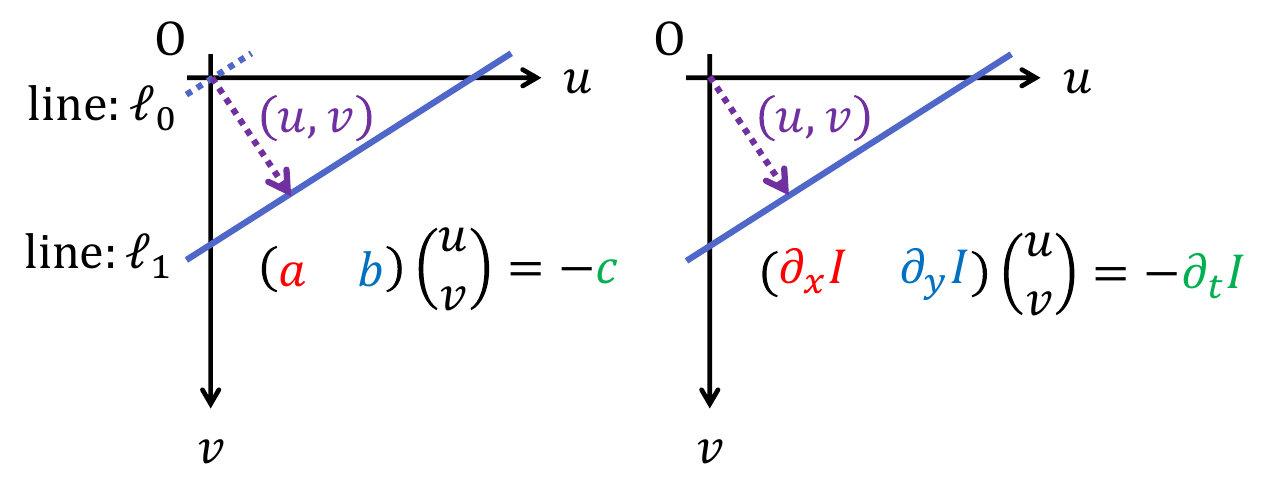}
    \caption{Relationship between edge-line translation and optical flow constraint for $(u,v)^\top$ at $\bm{x} \in (\ell_0\cap R)$: (left) constraint line from edge-line translation, (right) constraint line from brightness constancy equation.}
    \label{fig:gradient_plane}
\end{figure}

For each space-time sub-region $R \times [t_0, t_1]$ with $R \subset \Omega$, where $\Omega \subset \mathbb{R}^2$ is the image domain, approximating the event point cloud with a plane $ax + by + ct + d = 0$ yields two intersection lines, $\ell_0$ at $t_0$ and $\ell_1$ at $t_1$.
These two lines are related by translation, and their motion can be represented on the solution space of $u$ and $v$ as shown in the left panel of Fig.~\ref{fig:gradient_plane}.
The displacement between these lines divided by the time interval corresponds to the target flow velocity $\bm{u}(\bm{x},t_0) = (u(x,y,t_0),v(x,y,t_0))^\top$ for $\bm{x}\in(\ell_0\cap R)$. The relationship between the coefficients $a(\bm{x},t_0), b(\bm{x},t_0), c(\bm{x},t_0)$ of the plane equation and the flow components $u(\bm{x},t_0), v(\bm{x},t_0)$ is given by
\begin{equation}
a(\bm{x},t_0)u(\bm{x},t_0) + b(\bm{x},t_0)v(\bm{x},t_0) = -c(\bm{x},t_0).  \label{eq:plane}
\end{equation}
Here, within each region $R$, the functions $a(\bm{x},t_0),b(\bm{x},t_0)$, and $c(\bm{x},t_0)$ are constant on the line $\ell_0$ and zero elsewhere.

Similarly, for a frame image $I(\bm{x},t)$, the brightness constancy constraint equation based on the assumption of constant intensity over an infinitesimal time interval is expressed as
\begin{equation}
u(\bm{x},t) \partial_x I(\bm{x},t) + v(\bm{x},t) \partial_y I(\bm{x},t) = -\partial_t I(\bm{x},t). \label{eq:ofc}
\end{equation}
From this, the solution for the flow $\bm{u}$ from the frame image lies on the line shown in the right panel of Fig.~\ref{fig:gradient_plane}.

Therefore, from the correspondence between Eqs.~(\ref{eq:plane}) and (\ref{eq:ofc}), the coefficient functions $(a,b,c)$ of the plane equation formed by the events in each sub-region $R\times[t_0,t_1]$ correspond to the spatiotemporal gradients $(\partial_x I, \partial_y I, \partial_t I)$ of the frame image brightness, respectively. 
Although the coefficients $(a,b,c)$ can be obtained from each polarity, the two polarities $\{+, -\}$ can be considered as a two-channel image.
Then, following the Horn--Schunck type method~\cite{horn1981determining}, the optical flow is estimated by minimizing an objective functional combining a data term~(\ref{eq:ofc}) and a spatial smoothness term, defined as
\begin{dmath}
    \int_\Omega \sum_i c_i (\bm{u}^\top\nabla I_i + \partial_t I_i)^2 + \alpha (|\nabla u|^2 + |\nabla v|^2)  \,\mathrm{d}x \,\mathrm{d}y, \label{eq:hs_multi}
\end{dmath}
where $i$ denotes color channels (i.e., two polarities), $c_i$ is a positive weight between channels, $\nabla=(\partial_x, \partial_y)^\top$ and $\alpha$ is a positive value corresponding to the weight between terms~\cite{9287861}. The flow $\bm{u}$ minimizing \eqref{eq:hs_multi} is defined as the estimated flow.

\section{Proposed Method} \label{sec:proposed}
To capture temporal velocity variations, 
we extend the conventional method~\cite{10.1117/1.JEI.34.4.043024} along the temporal axis and propose a flow estimation method that accounts for acceleration.

\subsection{Temporal Division and Two-Step Planar Approximation}
\begin{figure}[t]
    \centering
    \includegraphics[width=\linewidth]{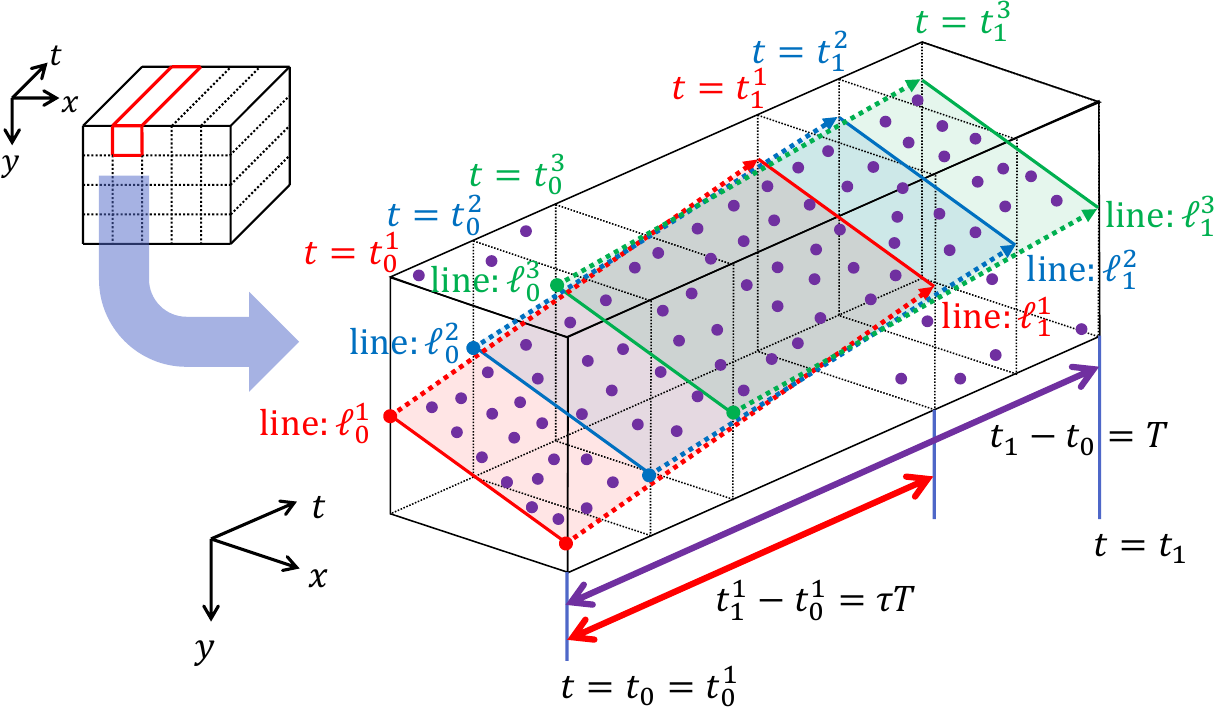}
    \caption{Overview of the proposed temporal division and sub-interval planar approximation.}
    \label{fig:proposed_concept}
\end{figure}

\begin{figure}
    \centering
    \includegraphics[width=0.9\linewidth]{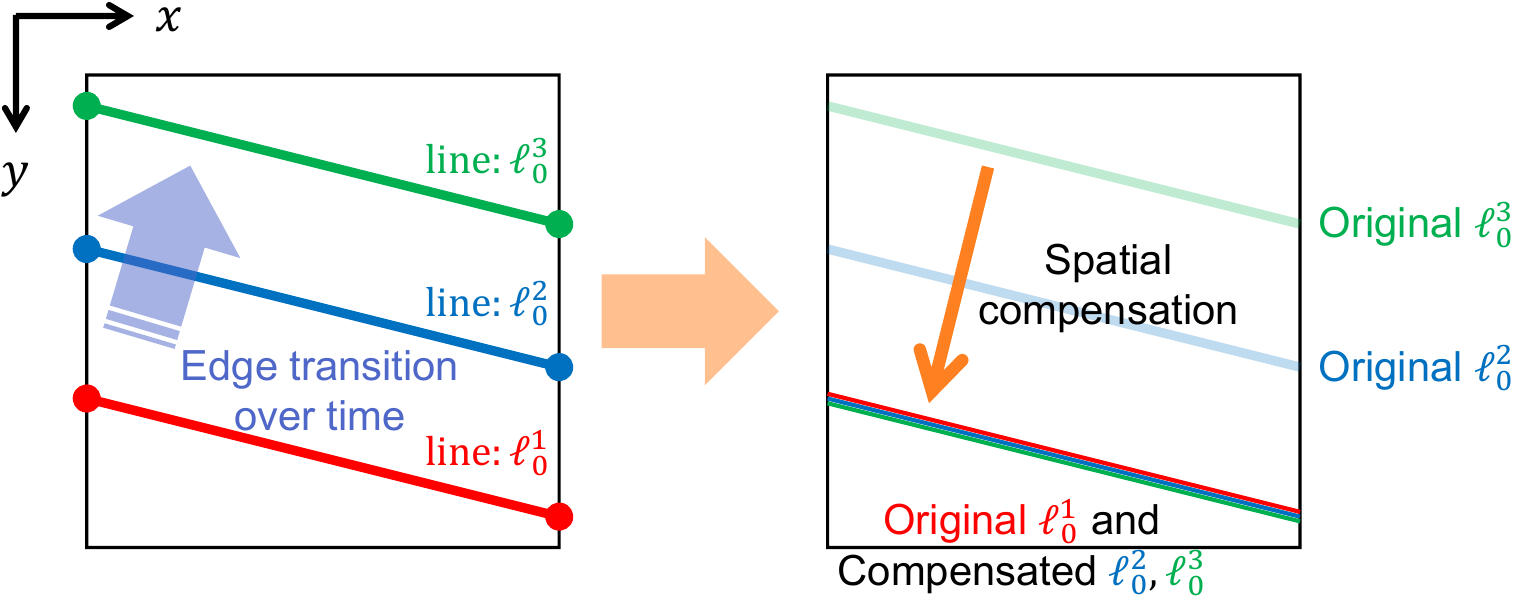}
    \caption{Spatial compensation of edge transitions of Fig.~\ref{fig:proposed_concept}. Gradients from sub-intervals (left) are spatially compensated to the reference position $\ell_0^1$ (right). This aggregates the spatiotemporal gradients of the same object at identical coordinates to maintain spatiotemporal consistency.}
    \label{fig:spatial_compensation}
\end{figure}

As illustrated in Fig.~\ref{fig:proposed_concept}, the proposed method divides the total estimation interval $[t_0, t_1]$ into $m$ sub-intervals $[t_0^j, t_1^j]$ ($j \in\{1,2,\dots,m\}$). For each sub-interval, the enclosed events are extracted, and spatiotemporal planar approximation is performed to estimate the local optical flow, thereby capturing temporal variations of the event surfaces. The resulting velocity field is then associated with the central timestamp $t_c^j = (t_0^j + t_1^j)/2$ of each sub-interval.

If planar approximation is performed independently within each short sub-interval, there is a risk of detecting different planes (motions) between consecutive intervals due to noise or interference from adjacent surfaces.
To mitigate this, the proposed method applies a two-step planar approximation: global motion is first identified by performing a planar approximation over the entire sequence, and detailed planar approximation for each sub-interval is then conducted using only the inliers obtained from this global approximation. This ensures consistent tracking of the same object edge across different sub-intervals.

Furthermore, if the gradient information obtained from each interval is used directly, spatial misalignment occurs as the object moves over time as shown in Fig.~\ref{fig:spatial_compensation}. Therefore, the edge position is determined based on the planar approximation result at a reference time, and the gradient information from each sub-interval is aggregated into this fixed spatial position. When computing the forward flow, the reference time is set to $t_0$. Here, forward flow denotes the displacement field from the reference time $t_0$ to $t_1$. This process compensates for spatial displacement over time, maintaining spatiotemporal consistency as the gradient information of the same object appears at identical coordinates across all intervals.

\subsection{Formulation with Temporal Regularization}
To estimate the flow, the obtained gradients are applied to an objective function. We modify \eqref{eq:hs_multi} by adding a temporal regularization term multiplied by a weight $\beta > 0$:
\begin{multline}
\int_{t_0}^{t_1}\int_\Omega \sum_{i} c_i (\bm{u}^\top\nabla I_i + \partial_t I_i)^2 + \alpha \left(|\nabla u|^2 + |\nabla v|^2\right) \\
+ \beta \left( (\partial_t u)^2 + (\partial_t v)^2 \right)  \,\mathrm{d}x\,\mathrm{d}y\,\mathrm{d}t.
\label{eq:hs_multi_time}
\end{multline}
The flow $\bm{u}$ that minimizes the objective function \eqref{eq:hs_multi_time} can be derived from the corresponding Euler--Lagrange equations, which, setting $c_i = 1$ for simplicity, are given by:
\begin{equation}
\alpha \nabla^\top \nabla \bm{u} + \beta \partial_{tt} \bm{u} = \sum_{i} (\bm{u}^\top\nabla I_i + \partial_t I_i) \nabla I_i. \label{eq:euler-lagrange}
\end{equation}

By introducing the spatial neighborhood average $\bar{\bm{u}}_s$ and the temporal neighborhood average $\bar{\bm{u}}_t$, \eqref{eq:euler-lagrange} can be approximated as follows:
\begin{equation}
\alpha' (\bar{\bm{u}}_s - \bm{u}) + \beta' (\bar{\bm{u}}_t - \bm{u}) = \sum_{i} (\bm{u}^\top\nabla I_i + \partial_t I_i) \nabla I_i, \label{eq:euler-lagrange_kinji}
\end{equation}
where $\bar{\bm{u}}_s$ is the weighted average of the 8 neighboring pixels at the same time~\cite{horn1981determining}, and $\bar{\bm{u}}_t$ is the average of the flow vectors at the same pixel between the preceding and succeeding intervals, following the spatial compensation described in Fig.~\ref{fig:spatial_compensation}.
The parameters $\alpha'=3\alpha/h^2$ and $\beta'=2\beta/\Delta t^2$ are coefficients associated with the space and time step sizes $h$ and $\Delta t$, respectively.
This discretization approach, which utilizes both spatial and temporal smoothness, can be viewed as an extension of the spatiotemporal Horn--Schunck framework~\cite{10.1007/BFb0014860,Weickert2001} applied to our event-based planar approximation.

To solve \eqref{eq:euler-lagrange_kinji}, the flow is obtained by iteratively applying the following update equations:
\begin{align}
u^{(k+1)} &= \frac{\alpha' \bar{u}_s^{(k)} + \beta' \bar{u}_t^{(k)} - \sum_{i} \partial_x I_i (v^{(k)} \partial_y I_i + \partial_t I_i)}{\alpha' + \beta' + \sum_{i} (\partial_x I_i)^2}, \label{eq:update_u}\\
v^{(k+1)} &= \frac{\alpha' \bar{v}_s^{(k)} + \beta' \bar{v}_t^{(k)} - \sum_{i} \partial_y I_i (u^{(k)} \partial_x I_i + \partial_t I_i)}{\alpha' + \beta' + \sum_{i} (\partial_y I_i)^2}, \label{eq:update_v}
\end{align}
where $u^{(k)}$ and $v^{(k)}$ denote the computed results after the $k$-th iteration.

Specifically, a straightforward pixel-wise linear regression is applied along the temporal axis to the time-series velocity fields $\{\bm{u}(\bm{x},t_c^1), \bm{u}(\bm{x},t_c^2), \cdots, \bm{u}(\bm{x},t_c^m)\}$ to estimate the initial velocity field $\bm{u}_0(\bm{x},t_0)$ and the mean apparent acceleration $\bm{g}(\bm{x},t_0)$ over the interval $T=t_1-t_0$. The forward flow is then obtained as the displacement from $t_0$ to $t_1$:
\begin{equation}
\bm{U}_{\mathrm{est}}(\bm{x},t_0)
=
\bm{u}_0(\bm{x},t_0)T
+(1/2)
\bm{g}(\bm{x},t_0)T^2.
\end{equation}

\section{Experiments}

\begin{table*}[t]
    \caption{Optical Flow Estimation Results on the DSEC Dataset~\cite{9387069} (AEE [pixel] $\downarrow$)}
    \label{tab:ablation_results}
    \centering
    \begin{threeparttable}
        \begin{tabular}{l cccc cccc cccc}
            \toprule
            & \multicolumn{4}{c}{\texttt{zurich\_{}city\_{}01\_{}a}} & \multicolumn{4}{c}{\texttt{zurich\_{}city\_{}02\_{}a}} & \multicolumn{4}{c}{\texttt{thun\_{}00\_{}a}} \\
            \cmidrule(lr){2-5}\cmidrule(lr){6-9}\cmidrule(l){10-13}
            Method & All & s-0-10 & s-10-40 & s-40+ & All & s-0-10 & s-10-40 & s-40+ & All & s-0-10 & s-10-40 & s-40+ \\
            \midrule
            Shiba \textit{et al.} (Flow, $L_1$)~\cite{10517639}\tnote{a} &
                                6.17 & 3.16 & 10.64 & 41.76 &
                                7.60 & 5.75 & 10.57 & 35.81 & 
                                3.73 & 3.02 & 4.96 & 27.40 \\
            Frame Flow\tnote{b}      & 
                                8.37 & 3.90 & 15.28 & 46.16 & 
                                14.87 & 5.07 & 18.34 & 44.89 & 
                                7.89 & 3.40 & 16.67 & 44.62 \\
            \midrule
            Baseline~\cite{10.1117/1.JEI.34.4.043024} &
                                3.38 & 2.27 & 4.84 & 20.55 & 
                                4.24 & 2.37 & 4.85 & 22.44 & 
                                4.10 & 2.45 & 7.32 & 28.60 \\
            +Time Div.               & 
                                3.14 & 2.24 & 4.24 & 19.43 & 
                                3.97 & 2.32 & 4.38 & 21.23 & 
                                3.53 & 2.19 & 6.16 & 27.15 \\
            +Spatial Comp.           & 
                                3.06 & 2.29 & 3.93 & 19.66 & 
                                3.96 & 2.27 & 4.26 & 21.83 & 
                                3.40 & 2.20 & 5.74 & 26.93 \\
            +Temporal Reg. (Proposed) & 
                                3.07 & 2.31 & 3.94 & 19.27 &
                                3.99 & 2.30 & 4.29 & 21.71 & 
                                3.37 & 2.20 & 5.63 & 26.20 \\
            \bottomrule
        \end{tabular}
        \begin{tablenotes}
            \item[a] Results are based on our custom implementation for the DSEC dataset~\cite{9387069}, referencing the GitHub code~\cite{GitHubtu78:online} and parameters in \cite{10517639}.
            \item[b] Evaluated by applying the energy functional (\ref{eq:hs_multi}) directly to frame images. Note that the AEE contains parallax errors caused by the physical baseline between the event and frame cameras.
        \end{tablenotes}
    \end{threeparttable}
\end{table*}

\newlength{\figwidth}
\setlength{\figwidth}{0.123\textwidth}

\begin{figure*}[t!]
\centering%
\hfill%
\begin{minipage}[b]{\figwidth}\centering\scriptsize Frame\end{minipage}%
\hfill%
\begin{minipage}[b]{\figwidth}\centering\scriptsize Events\end{minipage}%
\hfill%
\begin{minipage}[b]{\figwidth}\centering\scriptsize GT Flow\end{minipage}%
\hfill%
\begin{minipage}[b]{\figwidth}\centering\scriptsize GT Accel.\end{minipage}%
\hfill%
\begin{minipage}[b]{\figwidth}\centering\scriptsize Baseline Flow\end{minipage}%
\hfill%
\begin{minipage}[b]{\figwidth}\centering\scriptsize Estimated Flow\end{minipage}%
\hfill%
\begin{minipage}[b]{\figwidth}\centering\scriptsize Estimated Accel.\end{minipage}%
\hfill\mbox{}%
\\\vspace{1mm}
\hfill%
\begin{minipage}[b]{\figwidth}
    \includegraphics[width=\columnwidth]{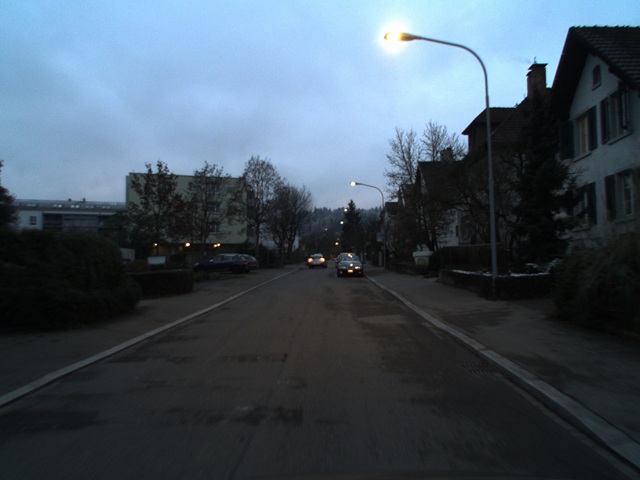}
\end{minipage}%
\hfill%
\begin{minipage}[b]{\figwidth}
    \includegraphics[width=\columnwidth]{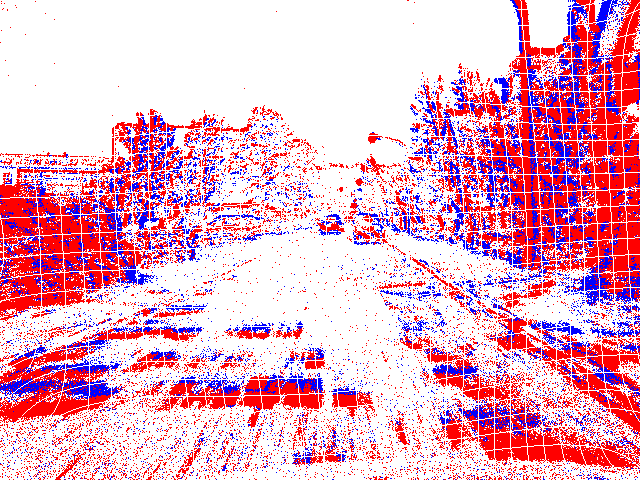}
\end{minipage}%
\hfill%
\begin{minipage}[b]{\figwidth}
    \includegraphics[width=\columnwidth]{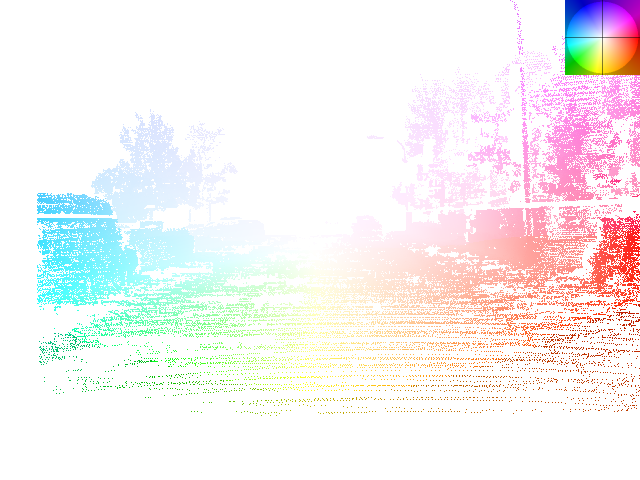}
\end{minipage}%
\hfill%
\begin{minipage}[b]{\figwidth}
    \includegraphics[width=\columnwidth]{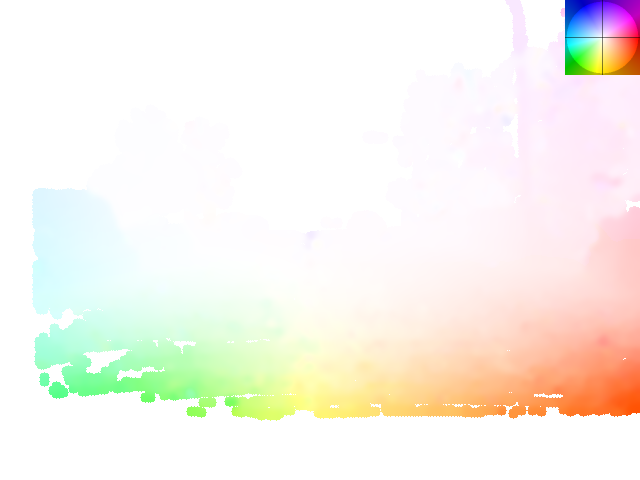} 
\end{minipage}%
\hfill%
\begin{minipage}[b]{\figwidth}
    \includegraphics[width=\columnwidth]{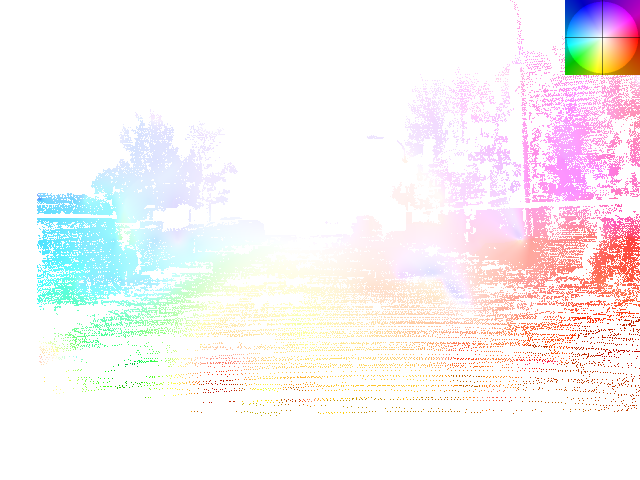} 
\end{minipage}%
\hfill%
\begin{minipage}[b]{\figwidth}
    \includegraphics[width=\columnwidth]{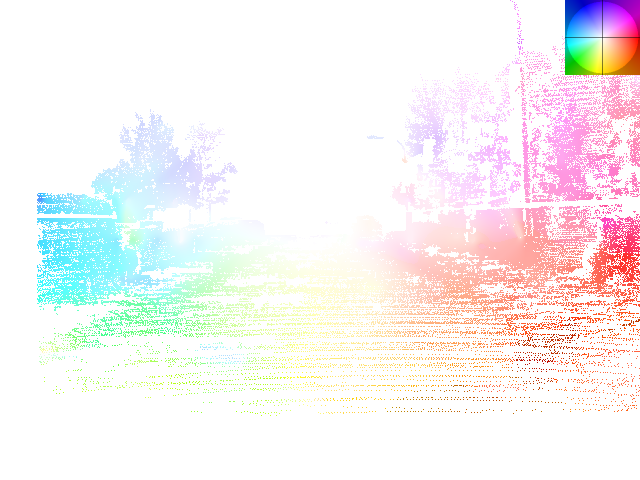} 
\end{minipage}%
\hfill%
\begin{minipage}[b]{\figwidth}
    \includegraphics[width=\columnwidth]{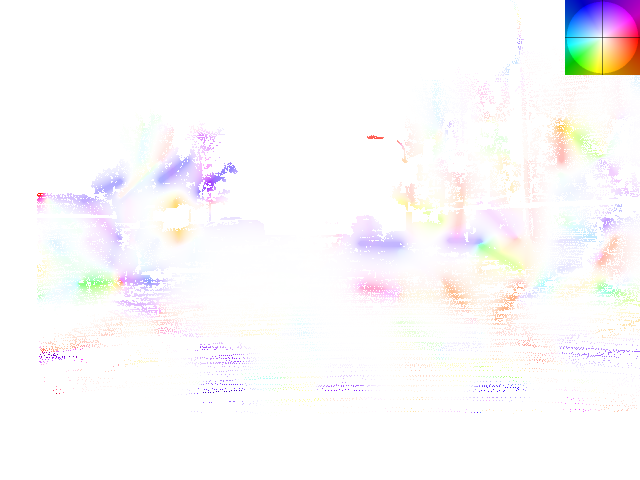} 
\end{minipage}%
\hfill\mbox{}%
\\\vspace{0.5mm}
\hfill%
\begin{minipage}[b]{\figwidth}
    \includegraphics[width=\columnwidth]{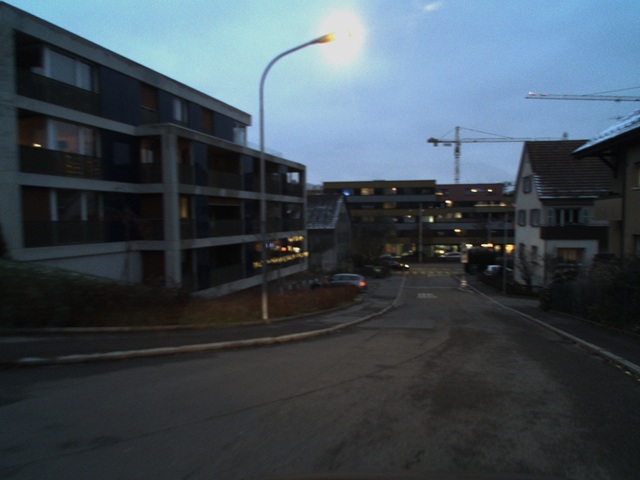}
\end{minipage}%
\hfill%
\begin{minipage}[b]{\figwidth}
    \includegraphics[width=\columnwidth]{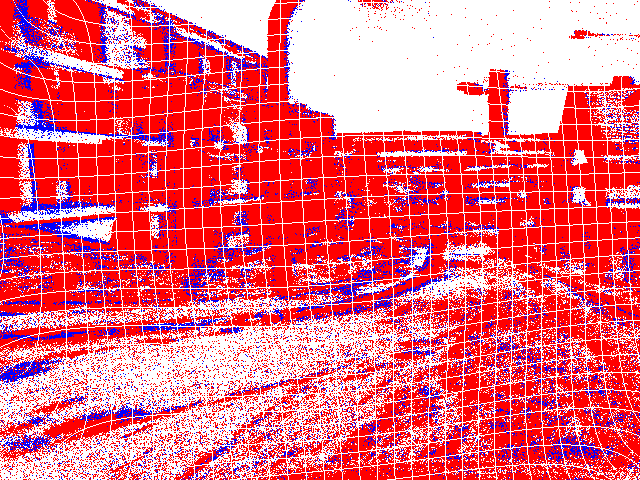}
\end{minipage}%
\hfill%
\begin{minipage}[b]{\figwidth}
    \includegraphics[width=\columnwidth]{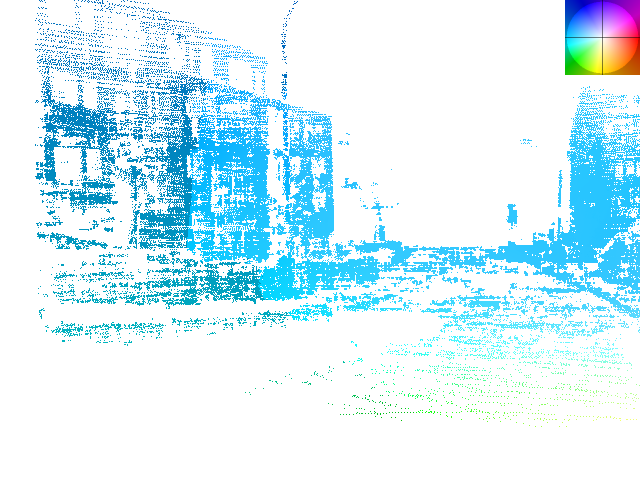}
\end{minipage}%
\hfill%
\begin{minipage}[b]{\figwidth}
    \includegraphics[width=\columnwidth]{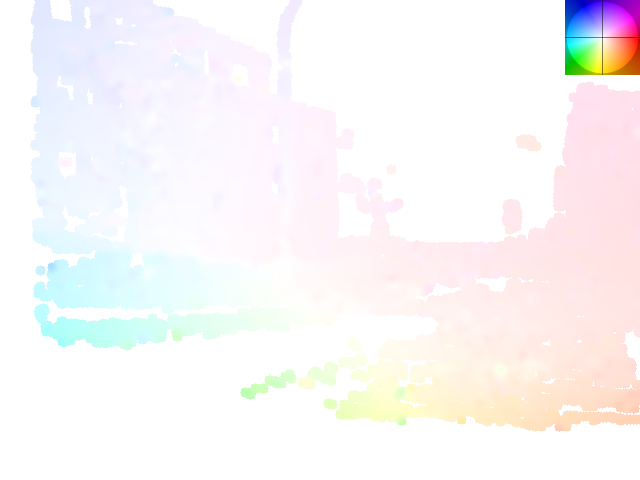} 
\end{minipage}%
\hfill%
\begin{minipage}[b]{\figwidth}
    \includegraphics[width=\columnwidth]{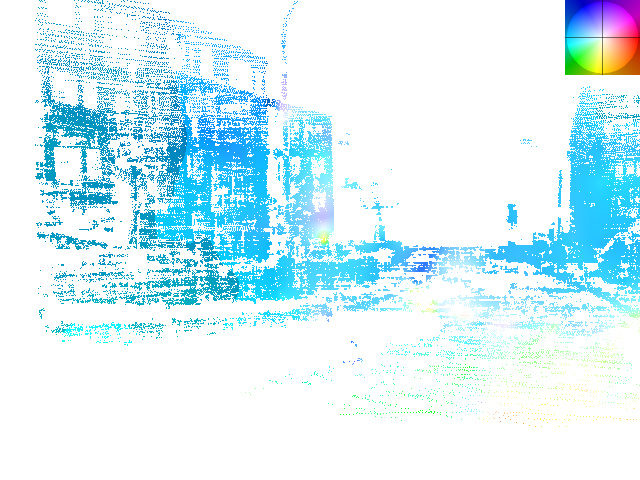} 
\end{minipage}%
\hfill%
\begin{minipage}[b]{\figwidth}
    \includegraphics[width=\columnwidth]{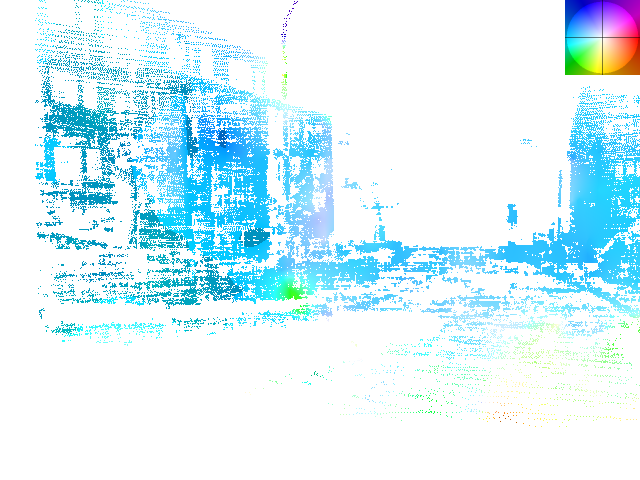} 
\end{minipage}%
\hfill%
\begin{minipage}[b]{\figwidth}
    \includegraphics[width=\columnwidth]{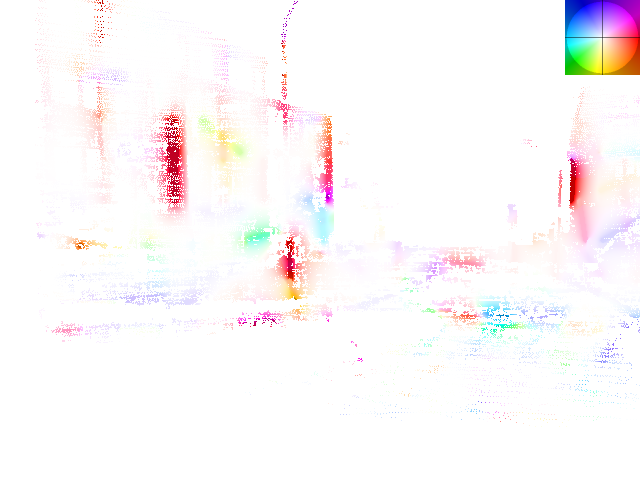} 
\end{minipage}%
\hfill\mbox{}%
\\\vspace{0.5mm}
\hfill%
\begin{minipage}[b]{\figwidth}
    \includegraphics[width=\columnwidth]{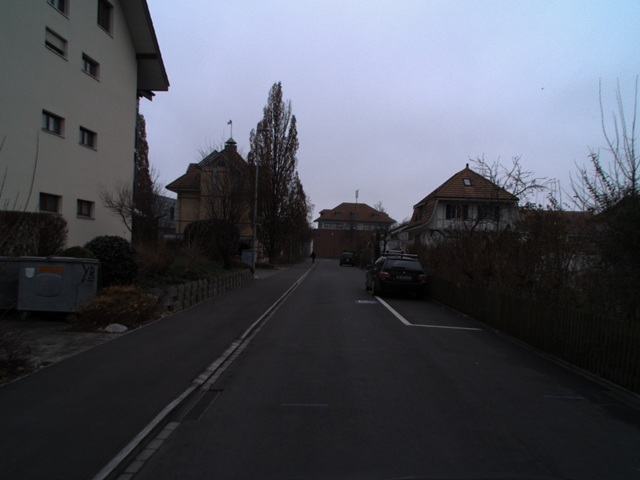}
\end{minipage}%
\hfill%
\begin{minipage}[b]{\figwidth}
    \includegraphics[width=\columnwidth]{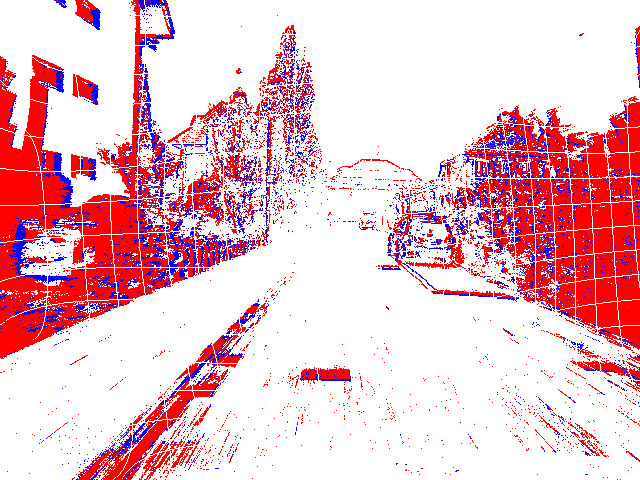}
\end{minipage}%
\hfill%
\begin{minipage}[b]{\figwidth}
    \includegraphics[width=\columnwidth]{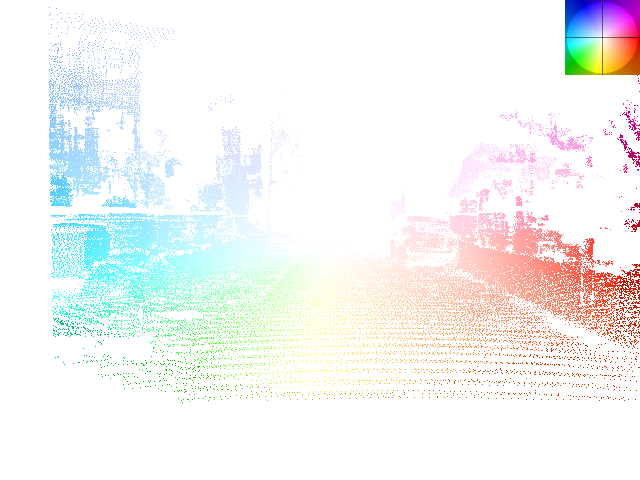}
\end{minipage}%
\hfill%
\begin{minipage}[b]{\figwidth}
    \includegraphics[width=\columnwidth]{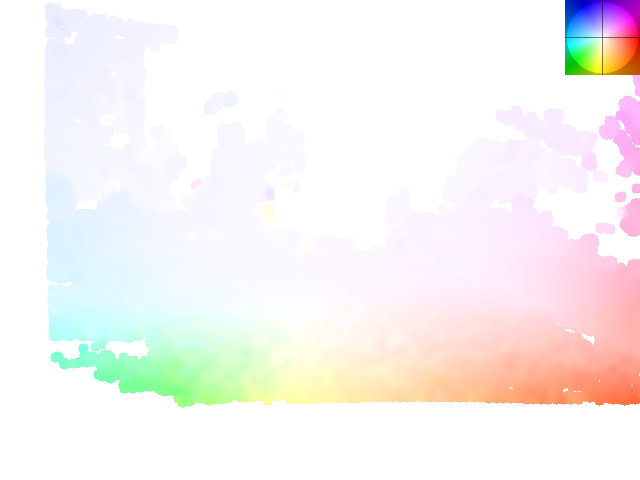} 
\end{minipage}%
\hfill%
\begin{minipage}[b]{\figwidth}
    \includegraphics[width=\columnwidth]{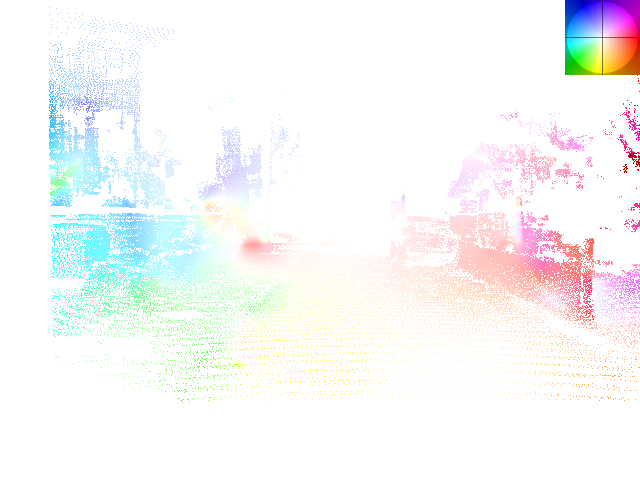} 
\end{minipage}%
\hfill%
\begin{minipage}[b]{\figwidth}
    \includegraphics[width=\columnwidth]{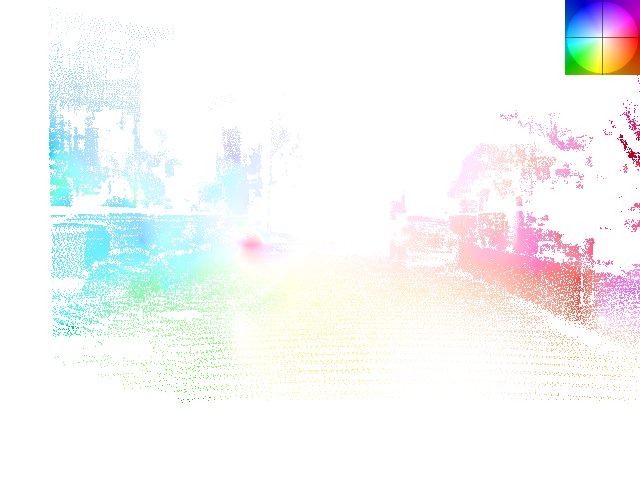} 
\end{minipage}%
\hfill%
\begin{minipage}[b]{\figwidth}
    \includegraphics[width=\columnwidth]{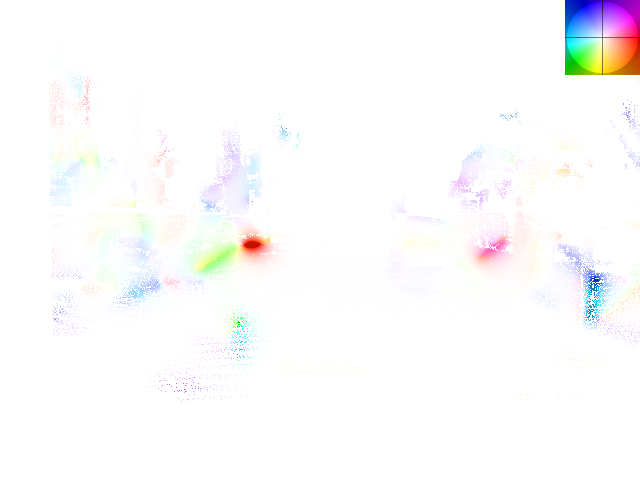} 
\end{minipage}%
\hfill\mbox{}%
\\
\caption{Visual comparison of results. Rows from top to bottom correspond to the \texttt{zurich\_city\_01\_a}, \texttt{zurich\_city\_02\_a}, and \texttt{thun\_00\_a} sequences.
Columns from left to right: Frame Image, Event Frame, Ground Truth (GT) Flow, GT Acceleration, Baseline Flow (without temporal division), Estimated Flow and Acceleration. Both estimated results are obtained by the proposed method with $m=5$ temporal divisions. The acceleration maps are shown only to illustrate the temporal variation captured by the proposed model; the primary evaluation criterion is the forward-flow AEE.}
\label{fig:visual_results_combined}
\end{figure*}

\subsection{Experimental Setup}
We evaluated the method on training sequences of the DSEC dataset~\cite{9387069}, which contains real-world automotive data, by computing the flow for each \qty{100}{ms} interval (i.e., $t_1-t_0=\qty{100}{ms}$), where ground-truth forward flow is available.

In the experiment, the VGA images ($640 \times 480$ pixels) were divided into small regions of $40 \times 40$ pixels, and the two-step planar approximation described in Sec.~\ref{sec:proposed} was applied, with the number of temporal divisions set to $m = 5$.
Specifically, using inliers extracted from the entire \qty{100}{ms} interval, sub-intervals spanning a fraction $\tau \coloneqq 1/2$ of the total length $T$ were arranged with temporal overlap and a detailed planar approximation was performed.
We set $m$ and $\tau$ empirically. Since the computational cost grows roughly linearly with $m$ whereas a larger $m$ is not expected to improve accuracy substantially, we considered $m = 5$ sufficient. The parameter $\tau$ balances temporal resolution against data sparsity.

Then, from the spatiotemporal gradients obtained for each interval, the flow was computed by solving \eqref{eq:euler-lagrange_kinji} under the spatiotemporal Neumann boundary condition. Notably, we replaced the standard RANSAC~\cite{10.1145/358669.358692} used in the previous study~\cite{10.1117/1.JEI.34.4.043024} with MSAC (M-estimator Sample Consensus)~\cite{710798,TORR2000138} and optimized some parameters. The inlier thresholds for the 1st and 2nd MSAC were set to 3.0 and 1.0, respectively. The objective function weights were set to $\alpha' = 0.5$ and $\beta' = 0.01$, where $\alpha'$ is relatively large to compensate for data sparsity, while $\beta'$ is small but nonzero, balancing temporal stability against over-smoothing of motion dynamics.
Optimization was performed for 5000 iterations initialized from zero velocity.

We used the Average Endpoint Error (AEE) [pixel] as the evaluation metric, defined as the $L_2$ norm of the error vector between the estimated flow $\bm{U}_{\mathrm{est}}$ and the ground-truth flow $\bm{U}_{\mathrm{gt}}$, i.e., $\|\bm{U}_{\mathrm{est}} - \bm{U}_{\mathrm{gt}}\|_2$, averaged over pixels with available ground-truth. Ground-truth acceleration was approximated from temporal differences of adjacent ground-truth flows.

\subsection{Evaluation Results and Ablation Analysis}
Table~\ref{tab:ablation_results} presents the optical flow estimation results on three DSEC~\cite{9387069} sequences. The proposed method surpasses the baseline~\cite{10.1117/1.JEI.34.4.043024} overall and in most motion ranges, and also shows competitive performance against \cite{10517639}. Note that our implemented baseline already incorporates improvements over the original---replacing standard RANSAC with MSAC and optimizing parameters---so the baseline itself already surpasses the scores reported in \cite{10.1117/1.JEI.34.4.043024}.

The ablation study clarifies the role of each component. Temporal division (+Time Div.) consistently reduces AEE, with larger gains in moderate and high speed ranges, confirming that relaxing the constant-velocity assumption is broadly effective regardless of motion magnitude. Spatial compensation (+Spatial Comp.) contributes to improving accuracy primarily in moderate-motion regions (s-10-40). Furthermore, the addition of temporal regularization (+Temporal Reg.) is effective for high-motion regions (s-40+). By temporally stabilizing the velocity estimation for large motions, which inherently tend to become unstable, it helps improve accuracy in the s-40+ range across all sequences. Meanwhile, the degradation in slow-motion regions (s-0-10) is relatively minor.

As visualized in Fig.~\ref{fig:visual_results_combined}, the estimated forward flow is qualitatively consistent with the ground truth. Meanwhile, despite challenges such as event sparsity and the difficulty of local plane fitting, the proposed method outputs apparent acceleration estimates with discernible differences in color distributions among sequences, indicating that motion orientations are roughly captured. Further improving the accuracy of acceleration estimation remains future work.

\section{Conclusion}
In this paper, we proposed an event-based optical flow estimation method that improves estimation accuracy by accounting for temporal velocity variations.
By partitioning the space-time volume into sub-intervals and applying spatial compensation to maintain consistency, our method improved the accuracy over the conventional baseline.
Temporal regularization further stabilizes the estimation, especially in high-motion regions, although its effect on AEE is sequence-dependent.
While the computational complexity scales roughly linearly with the number of sub-intervals $m$, reducing runtime through parallel processing and coarse-to-fine schemes remains an important direction for future work. Other future directions include comprehensive benchmarking against \cite{10517639} and learning-based methods, and addressing event sparsity through more advanced surface approximation techniques.

\bibliographystyle{IEEEtran}
\bibliography{refs}

\end{document}